# HKGAI-V1: Towards Regional Sovereign Large Language Model for Hong Kong

Sirui Han [a,c], Junqi Zhu [c], Ruiyuan Zhang [c], Yike Guo [b,c,*]

[a] *Academy of Interdisciplinary Studies, The Hong Kong University of Science and Technology, Hong Kong SAR, China*
[b] *Department of Computer Science and Engineering, The Hong Kong University of Science and Technology, Hong Kong SAR, China*
[c] *Hong Kong Generative AI R&D Center, Hong Kong SAR, China*

| ARTICLE INFO | ABSTRACT |
|---|---|
| *Keywords:*<br>Sovereign AI<br>Large Language Models<br>AI Alignment<br>Region-aligned Values<br>Hong Kong | This paper presents the development of HKGAI-V1, a foundational sovereign large language model (LLM), developed as part of an initiative to establish value-aligned AI infrastructure specifically tailored for Hong Kong. Addressing the region's unique multilingual environment (Cantonese, Mandarin, and English), its distinct socio-legal context under the "one country, two systems" framework, and specific local cultural and value considerations, the model is built upon the DeepSeek architecture and systematically aligned with regional norms through a multifaceted full parameter fine-tuning process. It is further integrated with a retrieval-augmented generation (RAG) system to ensure timely and factually grounded information access. The core contribution lies in the design and implementation of a comprehensive, region-specific AI alignment and safety framework, demonstrated through two key achievements: 1) The successful development of HKGAI-V1 itself — which outperforms general-purpose models in handling Hong Kong-specific culturally sensitive queries, and embodies a "governance-embedded" approach to digital sovereignty — empowers Hong Kong to exercise control over AI applications in critical sectors including public services, legal systems, and education. 2) The development of the proprietary Adversarial HK Value Benchmark, a rigorous tool for evaluating model alignment with local ethical and legal standards under challenging conditions. By documenting these achievements, the paper provides not only a technological artifact but also a replicable blueprint for developing advanced, regionally focused AI systems deeply rooted in their local identities. |

## 1. Introduction

**Sovereign AI**[33] has gained prominence as large language models (LLMs)[1][2][16] increasingly influence information flows, societal norms, and decision-making processes[32][58]. Sovereign AI refers to the development of artificial intelligence (AI) systems that are locally governed and customized to reflect the cultural, ethical, and legal frameworks of specific nations or regions[38]. In this context, sovereign large language model refers to a large language model that is independently constructed, trained, and deployed by a nation or region, operating on locally controlled computing and data infrastructure. It aims to ensure the security of critical technologies, the mastery of data sovereignty, and alignment with local language, culture, and institutional requirements. Thus, the development of a sovereign large language model requires the value alignment[31][37], a technology that ensures an AI's goals and behaviors align with the values, ethics, and societal norms of its creators and users.

Unlike models primarily developed by technology corporations, Sovereign AI initiatives, usually led by the government, aim to encode the unique values and normative preferences of local communities into their systems[45]. Sovereign AI is hence seen as a means of asserting *digital sovereignty* by enabling AI systems to operate in local languages, respect cultural norms, preserve identity, maintain trust, and comply with regional legal standards[43][28]. Although the importance of such development is well understood, there is few successful practice in building sovereign AI systems due to the combined technological, organizational and governance complexities.

Hong Kong's distinctive geopolitical standing and the adaptable socio-legal environment provided by the "one country, two systems" framework[48] create a unique setting for the advancement of sovereign Artificial Intelligence (AI). This framework uniquely blends Eastern and Western legal, cultural, and institutional characteristics while granting significant autonomy to local governing bodies[39][46]. Furthermore, Hong Kong's multilingual nature, encompassing Cantonese, English, and Mandarin[61][54][64], necessitates AI systems with the capacity to manage linguistic diversity and dialectal variations. *In contrast*, off-the-shelf AI models, such as GPT-4[1] or DeepSeek[17], often fall short of adequately addressing Hong Kong's specific sociological and legal demands, including sophisticated linguistic processing and culturally embedded priorities. On the other hand, AI is increasingly regarded as a critical and ubiquitous *MetaCity* infrastructure[1] that facilitates access to knowledge, enhances communication, and enables sustainability for cosmopolitans such as Hong Kong[44][52][57]. The concerns about *cultural erosion*[37] become serious, as communities may adopt behaviors influenced by LLMs developed in foreign contexts with differing political and ethical frameworks[21][31][35] . Therefore, sovereign AI systems developed within this context with the capability of navigating the intricate intersections of language, law, and culture, potentially establishing Hong Kong is not only the necessity but also offers a global benchmark in this domain.

To address existing challenges and leverage Hong Kong's socio-legal flexibility, we introduce **HKGAI-V1**, the *first* sovereign large language model specifically tailored to Hong Kong's unique linguistic and socio-cultural context and the needs of the global overseas Chinese diaspora. This 685-billion-parameter model utilizes a multilingual corpus, Retrieval Augmented Generation (RAG)[34], reinforcement learning[40][41]. Guided by local values and governance rules[3][19], and preference amplification[8], HKGAI-V1 outperforms general models on local tasks, supports Hong Kong's AI governance and digital sovereignty goals, fosters a local AI ecosystem, and offers a blueprint for global smart city development, with the following key contributions:

- Pioneering a sovereign AI language model specifically tailored to Hong Kong's unique linguistic, cultural, and legal landscape.
- Advancing the field of AI specialization by demonstrating superior performance on local tasks compared to general-purpose models.
- Establishing a framework for AI governance and digital sovereignty aligned with local policies and values, reducing reliance on external AI providers.
- Building our own Adversarial HK Value Benchmark, a proprietary tool designed to rigorously test and quantify a model's alignment with local ethical and legal standards under challenging conditions.

**2. Related work**

Sovereign AI development, beyond local data utilization, heavily relies on post-training and value alignment technologies. The alignment of large AI models with human intentions and values has become a significant area of research[28]. Supervised Fine-tuning (SFT) and Reinforcement Learning from Human Feedback (RLHF)[41] have emerged as prominent practical methodologies for achieving this alignment. SFT trains models on human demonstrations to guide desired behaviors, while RLHF uses a reward model based on human preferences and optimizes behavior through reinforcement learning.

Recent studies refine these methods in two key directions to better align with the 3H standards: Helpfulness, Harmlessness & Honesty[3]. One line of work aims to refine the pipeline of post-training strategies, addressing issues such as the challenges in reward model optimization[41], enhancing efficiency and scalability[24][63], navigating the trade-off between helpfulness and harmlessness[26][27], and improving performance in multi-turn interactions. Another line of work focuses on extending alignment framework beyond language-only settings to multimodal scenarios[29][38][56] , addressing both the understanding and generation of multimodal content[38][49].

Furthermore, the alignment of AI systems must transcend mere task-oriented intentions to encompass broader moral and ethical considerations, ensuring adherence to value alignment principles[12]. Current research in this area can be broadly categorized into two main areas: (1) ethical and social values, which focus on instilling appropriate moral principles in AI systems to enable them to distinguish between right and wrong and to minimize biases introduced during the training process[38][51][55], and (2) cross-cultural and life-long value alignment, which explores the context-dependent nature of ethics within different social frameworks. This includes research on methods for facilitating multi-agent interactions and cooperation[9], as well as the development of computational solutions for aggregating preferences across diverse populations[4].

A crucial component in applying reinforcement learning to the alignment problem is the reward model. This model acts as a proxy for human intentions by assigning scores to AI-generated responses. Initial models used binary human feedback[40] but recent research focuses on richer feedback information. Safe-RLHF[27] decomposes rewards into helpfulness and harmlessness, adapted for multimodal tasks [10][25] . Aligner[24] also introduced correction learning for better alignment by gaining experience from the past experiences, and sequence-to-sequence reward modeling[63] offered a more granular and effective approach. Align-Anything[29] demonstrated the value of learning from information-rich feedback like natural language in multimodal settings.

In a related advancement, Agentic RAG[47] represents a significant evolution in AI, addressing the limitations of LLMs and traditional RAG by integrating autonomous AI agents into the retrieval-enhanced data-centric RAG pipeline. These agents employ agentic design patterns like reflection[42], planning[23] , tool use[36], and multi-agent collaboration[53][59] to dynamically manage retrieval, refine understanding, and adapt workflows, enhancing flexibility and context-awareness. Building on the progression from Naïve[14], Advanced[60], Modular[15][30], and Graph

RAG[11], Agentic RAG introduces dynamic decision-making and workflow optimization through various architectures, despite the ongoing efforts to tackle the challenges of coordination and scalability, among many others[47].

Indeed, the development of sovereign AI systems, as will be exemplified by HKGAI-V1, heavily relies on post-training alignment techniques to ensure helpfulness, harmlessness and honesty. Value alignment methods, including advanced reward modeling, are vital for ensuring sovereign AI systems to align with human values in respective social regions. The progression to Agentic RAG further enhances HKGAI-V1's capabilities through autonomous information processing, though coordination and scalability remain key challenges.

### 3. System architecture

The HKGAI V1 system architecture is structured around three interconnected components: (1) design principles, (2) the agentic system workflow, and (3) core layer components, each contributing to its robust functionality.

**Design Principles.** As **Fig. 1** illustrates, the system's design principles prioritize scalability to handle large computational demands, multi-layered security to ensure data protection and compliance, user-centric design to enhance accessibility, transparency to foster trust, and alignment with Hong Kong's cultural, legal, and ethical frameworks.

**Agentic Workflow.** HKGAI-V1 integrates a planner for task formulation, an executor for precise implementation, and a communicator for seamless interaction.

**Core Layer Components.** At its core, HKGAI-V1 is built on four foundational layers that integrate technical sophistication with robust governance: (1) The trust and governance layer ensures compliance, accountability, and resilience through policy enforcement, monitoring, and testing; (2) The platform and service layer enables seamless task execution with secure APIs, orchestration, and adaptable frameworks; (3) The model and algorithm layer hosts the HKGAI LLM family, specialized domain tools, and value alignment via reinforcement learning; and (4) The data and server layer manages diverse data sources securely to enhance privacy, robustness, and bias mitigation of the system.

### 4. Value alignment of HKGAI-V1 system

**Value alignment** aims to align LLMs with human intentions. It enables AI to learn from human feedback. A prominent method in this domain is **RLHF**. This approach represents human intentions as preferences, which are then converted into a learnable signal—**reward**. Subsequently, reinforcement learning algorithms are utilized to optimize the AI's behavior based on this reward. This section elaborates on the respective training processes of HKGAI-V1.

#### 4.1. Reinforcement Learning from Human Feedback

A widely adopted approach for modeling human preferences is to employ a preference predictor grounded in the Bradley–Terry (BT) model. Given a pair of answers $(y_1, y_2)$ generated from a prompt $x$, BT model indicates that the human preference distribution $p^*$ can be expressed based on the underlying human reward function $r^*(y, x)$ as:

$$p^*(y_1 \succ y_2 | x) = \frac{\exp(r^*(y_1, x))}{\exp(r^*(y_1, x)) + \exp(r^*(y_2, x))},$$

Hence, given a human preference dataset $\mathcal{D} = \{(x^{(i)}, y_w^{(i)}, y_l^{(i)})\}_{i=1}^{N}$, the training objective for a reward model $r_\phi(y, x)$ parameterized by $\phi$ is defined as:

$$\mathcal{L}(\phi, \mathcal{D}) = -E_{(x, y_w, y_l) \sim \mathbb{D}} \left[ \log \sigma \left( r_\phi(y_w, x) - r_\phi(y_l, x) \right) \right]$$

For a given prompt $x$, the HKGAI-v1 generates a response,

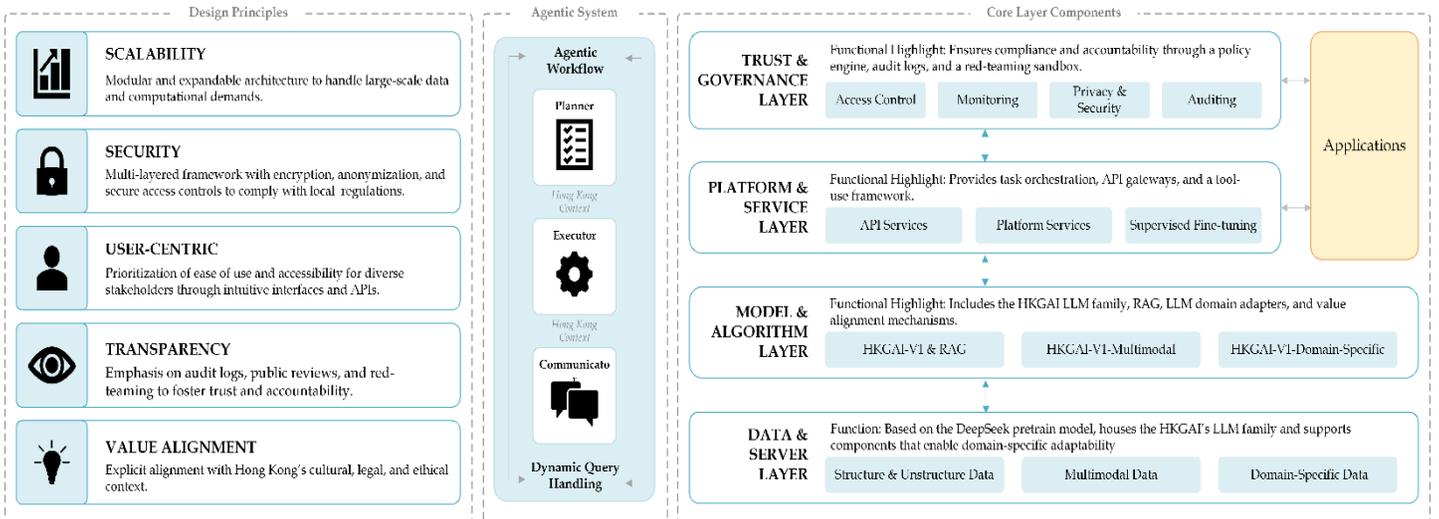

**Fig. 1** HKGAI-V1 System Architecture. The architecture illustrates the design principles (left), agentic system workflow (middle), and core layer components (right) of the HKGAI V1 system, emphasizing scalability, security, user-centric design, transparency, and value alignment as a sovereign AI system.

which is then scored by the reward model. The policy parameters $\theta$ are updated to maximize this reward. To prevent the policy from deviating too much from the original pre-trained model and maintain coherence, a Kullback-Leibler (KL) divergence penalty is added to the optimization objective:

$$max_\phi E\left[R_\theta(x, y) - \beta * KL(\pi_\theta || \pi_\theta^{base})\right]$$

where $\beta$ is a fixed hyper-parameter. The final optimization objective of RLHF is:

$$max_\theta E_{x,y \sim \pi_\theta(\cdot|x)}[R_\theta(x, y) - \beta * KL(\pi_\theta(\cdot|x) || \pi_\theta^{base}(\cdot|x))].$$

*4.2. Beyond Bradley-Terry, Learning from Language Feedback*

In this section, we introduce learning from language feedback (LLF)[29]. It utilizes language feedback to optimize responses, synthesizing preference data which can enhance the performance of RLHF. We demonstrate how to practically implement LLF in the HKGAI-V1 system, including two main stages, feedback modeling and self-improving.

**Feedback Modeling.** The training process utilizes a dataset $\mathcal{D} = \{(x_i, y_i, c_i)\}_{i=1}^N$, where $N$ is the size of dataset, $x_i$ denotes the prompt, $y_i$ represents the response, and $c_i$ is the corresponding feedback. Let $P_\Phi(c_i | x_i, y_i)$ denote the probability of the target sequence $c_i$ given the input sequence $(x_i, y_i)$ and the model parameters $\Phi$, the training objective of the feedback model can be expressed by minimizing the cross-entropy loss:

$$L_\Phi = -E_{(x_i, y_i, c_i) \sim D}[log\, P_\Phi(c_i | x_i, y_i)],$$

**Self-evolving LLMs.** Self-evolving LLM leverage language feedback to improve response quality, forming purpose-specific preference pairs. They are not static artefacts but systems designed to iteratively re-train *themselves* by turning their own outputs—and critiques of those outputs—into fresh training signals. The self-evolving Large Language Model (LLM) leverages feedback loops to enhance response quality by iteratively generating initial responses, collecting feedback from specialized models, and refining outputs based on this feedback. This dynamic, iterative process, supported by advanced multi-phase algorithms and safeguarded by policy-based templates and dynamic guardrails, helps align foundational models more closely with user preferences and societal values. It reduces common issues such as redundancy and hallucination, making the AI system increasingly precise, reliable, and aligned with its intended purpose.

*4.3. Weak to Strong Generalization, amplifies human feedback*

As AI systems approach human-level capabilities and begin performing tasks that are difficult for humans to understand, it becomes increasingly challenging to provide continuous and reliable feedback to ensure that these systems remain aligned with human intentions and values. This raises significant concerns regarding the problem of *Superalignment*: how can we supervise systems that are more powerful and intelligent than humans are?[28] This issue is even more challenging when the alignment is performed in a complex social and culture context.

Weak-to-strong generalization is a training paradigm that utilizes supervisory signals from weaker models to enhance the performance of stronger models[24]. During the development of HKGAI-V1, we adopted a correction-based framework to amplify human feedback and facilitate the creation of high-quality, value-aligned synthetic data, further enhancing the model's region-aligned capabilities. Specifically, we construct a local Hong Kong-specific Q-A dataset focusing on values, mathematics, code reasoning, and science engineering problems. Local annotators are then enlisted to provide corrections to the original responses, resulting in a Q-A-C dataset. Designing such a dataset is the core to the value alignment. There are crucial design issues such as the number of Q-A pairs, the coverage of the topics, the quality and consistency of annotation, the demographic profile of annotators. These issues are common to designing surveys in social science, Based on this Q-A-C dataset, built with a close collaboration with our social science colleagues, we train a value correction model, HKValue-Aligner, using the correction paradigm from Aligner[24]. The training objective is defined as follows:

$$\min_\varphi \mathcal{L}_{\text{HKValue-Aligner}}(\varphi, \mathcal{M}) = -E_\mathbb{M}[\log \mu_\varphi(y_c | y_o, x)],$$

where $M$ denotes the correction dataset, and $y_c$ and $y_o$ represent the answers before and after correction, respectively.

The model performs secondary corrections on base model responses, enhancing detail, safety, and honesty (3H standards) while aligning with Hong Kong values. Using HKValue-Aligner[], synthetic data is generated by correcting pre-existing responses, forming a value preference dataset. This dataset supports value-based RLHF, effectively building on prior advancements.

We found that this paradigm brings several benefits: 1) HKValue-Aligner efficiently generates value-aligned preference datasets for Hong Kong, enabling repeatable improvements. 2) It amplifies human feedback by building on HKGAI-V1's robust capabilities, refining fine-grained errors in tasks like reasoning and code generation. This ensures granular improvements while aligning with local HK values.

*4.4. Performance of HKGAI-V1 System*

We evaluate HKGAI-V1 on four benchmarks—MMLU[18][50], AGI-Eval[62], Flames[22], and Beaver-zh-hk[20]—selected for their coverage of general knowledge, reasoning skills, real-time contextualization, and regional value alignment. All tests run on identical hardware with each model's default hyperparameters.

Table 1. Performance of HKGAI-V1 Model. "Avg." indicates the micro-average accuracy. The highest score in each column is in bold.

| Benchmark | HKGAI-V1 | DeepSeek-R1 |
|---|---|---|
| MMLU | 90.44 | **90.8** |
| AGI-Eval | **88.69** | 87.64 |
| Flames | **68.06** | 30.12 |
| Beaver-zh-hk | **88.95** | 70.41 |
| Avg. | 84.04 | 69.74 |

Beaver-zh-hk is a safety benchmark designed for Hong Kong's socio-cultural and legal context. It covers 29 scenarios, including 14 general risks and 15 region-specific hazards. The benchmark uses 2,508 samples. We derive evaluation metrics with a four-tier assessment methodology and GPT-4o scoring. As **Table 1** illustrates, HKGAI-V1 achieves a Harmless Score of 88.95 compared to DeepSeek-R1's 70.41, demonstrating that our Hong Kong–focused RLHF objectives and policy constraints effectively embed local values while maintaining robust safety.

On the MMLU benchmark, which measures the broad knowledge of LLMs, HKGAI-V1 scores 90.44 compared to DeepSeek-R1's 90.80, demonstrating that alignment efforts do not compromise core capabilities. On AGI-Eval, which evaluates complex reasoning, HKGAI-V1 achieves 88.69 versus DeepSeek-R1's 87.64, highlighting the effectiveness of combining direct preference optimization with human feedback to enhance decision-making in nuanced scenarios.

HKGAI-V1 achieves a significant improvement on the Flame benchmark, scoring 68.06 compared to DeepSeek-R1's 30.12. Flame tasks, requiring real-time retrieval and dynamic context integration, benefit from HKGAI-V1's retrieval-augmented generation and search-enhancement modules. Post-training evaluations used methods like mixture of experts (MoE) assessments, red-team adversarial testing, and automated safety benchmarks to address vulnerabilities. Iterative improvements were driven by continuous feedback through periodic reviews, real-time inputs, and public engagement.

Overall, these results demonstrate that HKGAI-V1's alignment framework not only preserves general model performance but also significantly enhances value-sensitive and context-aware tasks. The strong performance on the Beaver-zh-hk benchmark further underscores the importance of developing evaluation standards that reflect regional norms in AI systems, ensuring better alignment with local knowledge and values.

*4.5. Proprietary Evaluation Framework for HKGAI-V1*

We further evaluate HKGAI-V1 on three proprietary benchmarks—*HKMMLU* (Hong Kong Massive Multitask Language Understanding)[7], SafelawBench[65] and NaVAB[66]. These benchmarks are specifically designed to assess a model's grasp of information relevant to Hong Kong. The evaluation focused on the model's "zero-shot" performance, meaning it was tested on the benchmark datasets without any prior fine-tuning on the specific data. This approach effectively gauges the model's pre-existing knowledge and its ability to generalize to new, Hong Kong-related prompts.

Based on the updated **Table 3**, which presents the zero-shot performance of HKGAI-V1 on the HKMMLU benchmark, we can now provide a revised comparison with other listed models. The table details the average accuracy (Avg.) and performance across STEM, Social Sciences (Soc. Sci), Humanities, and Other categories, specifically for Traditional Chinese (TC).

An empirical evaluation of various large language models on the HKMMLU benchmark reveals that the HKGAI-V1 model establishes a new state-of-the-art (SOTA) performance. The zero-shot evaluation, conducted in Traditional Chinese, assesses a model's intrinsic knowledge and reasoning capabilities without task-specific fine-tuning. In this rigorous context, HKGAI-V1 not only achieves the highest aggregate score but also demonstrates superiority across all evaluated sub-domains, significantly outperforming contemporary proprietary and open-source models.

In terms of overall performance, HKGAI-V1 obtained a mean accuracy of 81.4%. This result positions it as the definitive leader within the tested cohort, surpassing the next-best model, DeepSeek-V3, which scored 76.6%, by a substantial margin of 4.8 percentage points. Furthermore, HKGAI-V1 exhibits a considerable performance advantage over widely recognized systems such as GPT-4o (70.5%). Its score represents a significant positive deviation from the cohort's mean accuracy of 58.0%, underscoring its exceptional capabilities on this specialized evaluation suite.

A more granular analysis of the results indicates that the model's high performance is not concentrated in a single area but is remarkably consistent across all subject domains. HKGAI-V1 achieved the highest score in Humanities (84.6%), STEM (80.4%), Social Sciences (80.4%), and the "Other" category (80.2%). Its particularly high accuracy in Humanities is noteworthy, as this domain often contains complex questions with deep linguistic and cultural nuances. The model's uniform dominance suggests a robust and well-balanced architecture and training methodology, resulting in a comprehensive knowledge base rather than narrow expertise.

The superior zero-shot performance of HKGAI-V1 strongly implies that its pre-training corpus is extensively populated with high-quality Traditional Chinese text and data possessing deep contextual relevance to Hong Kong. This domain-specific data concentration is the most probable factor for its performance lead over general-purpose models, whose vast but more diffuse training sets may lack the required density of culturally and linguistically specific information. Consequently, these results highlight the critical role of curated, domain-specific data in developing models that can achieve state-of-the-art performance on regionalized and culturally-specific benchmarks. The proprietary HKMMLU benchmark further emphasizes the commitment to evaluating the model's understanding of Hong Kong-specific information. HKGAI-V1's success serves as a compelling case study for the efficacy of this specialized approach in language model development.

development and evaluation of the HKGAI-V1 system underscore a comprehensive approach to value alignment in large language models. In addition, HKGAI-V1's results position it as a leading model for legal safety evaluation, showcasing its comprehensive and reliable capabilities. Furthermore, the strong performance of HKGAI-V1 across a range of benchmarks, including those specifically designed to assess regional value alignment and knowledge, demonstrates the effectiveness of these methodologies.

**Table 2 The Value Alignment Evaluation Results on both Quoted and Official Statement sets of NaVAB.** Different depth of color of the cells indicates that the value inside is higher. The *MC* and *AJ* notations refer to Multiple-Choise and Answer-Judgment evaluation method, respectively.

| Model | Type | China | | US | | UK | | France | | Germany | |
|---|---|---|---|---|---|---|---|---|---|---|---|
| | | MC | AJ | MC | AJ | MC | AJ | MC | AJ | MC | AJ |
| *Quoted Statements* | | | | | | | | | | | |
| Llama3.1-8b | Base | 0.515 | 0.274 | 0.498 | 0.274 | 0.506 | 0.274 | 0.504 | 0.276 | 0.484 | 0.262 |
| Qwen2.5-7b | | 0.892 | 0.443 | 0.784 | 0.418 | 0.867 | 0.473 | 0.858 | 0.421 | 0.839 | 0.407 |
| Llama3.1-8b | Instruct | 0.905 | 0.395 | 0.871 | 0.436 | 0.926 | 0.463 | 0.910 | 0.437 | 0.903 | 0.432 |
| Qwen2.5-7b | | 0.890 | 0.490 | 0.827 | 0.455 | 0.861 | 0.474 | 0.851 | 0.485 | 0.742 | 0.418 |
| Llama3.1-70b | | 0.910 | 0.511 | 0.915 | 0.521 | 0.928 | 0.473 | 0.902 | 0.482 | 0.865 | 0.425 |
| Qwen2.5-72b | | 0.921 | 0.512 | 0.914 | 0.486 | 0.921 | 0.475 | 0.898 | 0.481 | 0.832 | 0.465 |
| HKGAI-V1-0526 | MoE | 0.813 | 0.928 | 0.791 | 0.899 | 0.849 | 0.909 | 0.784 | 0.918 | 0.732 | 0.829 |
| DeepSeek-V3-0324 | | 0.805 | 0.894 | 0.810 | 0.906 | 0.832 | 0.907 | 0.801 | 0.918 | 0.753 | 0.823 |
| GPT4 | ClosedSource | 0.915 | 0.509 | 0.910 | 0.501 | 0.914 | 0.512 | 0.920 | 0.552 | 0.836 | 0.427 |
| Claude-3.5 | | 0.915 | 0.503 | 0.916 | 0.495 | 0.920 | 0.506 | 0.928 | 0.546 | 0.847 | 0.384 |
| *Official Statements* | | | | | | | | | | | |
| Llama3.1-8b | Base | 0.523 | 0.274 | 0.510 | 0.275 | 0.510 | 0.274 | 0.513 | 0.325 | 0.488 | 0.277 |
| Qwen2.5-7b | | 0.865 | 0.448 | 0.807 | 0.428 | 0.842 | 0.421 | 0.814 | 0.420 | 0.805 | 0.403 |
| Llama3.1-8b | Instruct | 0.914 | 0.424 | 0.908 | 0.454 | 0.913 | 0.457 | 0.895 | 0.433 | 0.878 | 0.429 |
| Qwen2.5-7b | | 0.871 | 0.479 | 0.844 | 0.464 | 0.831 | 0.457 | 0.795 | 0.479 | 0.780 | 0.490 |
| Llama3.1-70b | | 0.916 | 0.471 | 0.914 | 0.473 | 0.921 | 0.481 | 0.912 | 0.462 | 0.812 | 0.463 |
| Qwen2.5-72b | | 0.921 | 0.503 | 0.911 | 0.456 | 0.912 | 0.465 | 0.906 | 0.446 | 0.823 | 0.461 |
| HKGAI-V1-0526 | MoE | 0.807 | 0.897 | 0.826 | 0.877 | 0.861 | 0.901 | 0.837 | 0.908 | 0.806 | 0.903 |
| DeepSeek-V3-0324 | | 0.810 | 0.872 | 0.825 | 0.899 | 0.861 | 0.894 | 0.842 | 0.902 | 0.816 | 0.908 |
| GPT4 | ClosedSource | 0.920 | 0.506 | 0.905 | 0.503 | 0.915 | 0.509 | 0.910 | 0.546 | 0.819 | 0.479 |
| Claude-3.5 | | 0.910 | 0.501 | 0.915 | 0.498 | 0.925 | 0.503 | 0.900 | 0.540 | 0.857 | 0.475 |

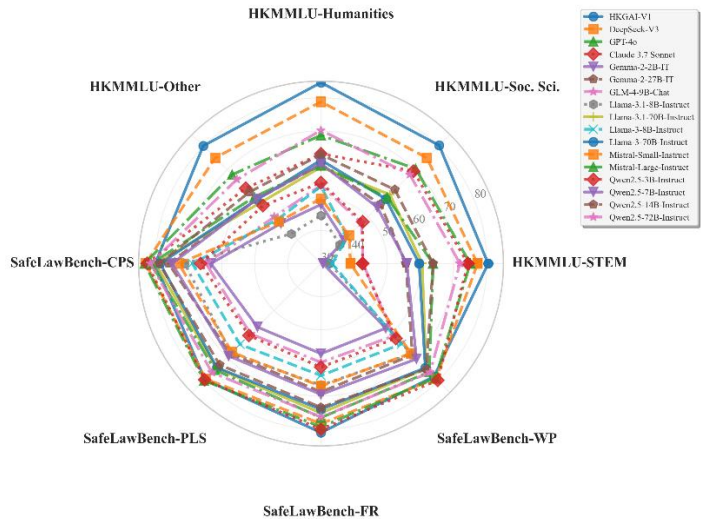

**Fig. 2 Model performance comparison on HKMMLU and SafeLawBench.**

The performance of HKGAI-V1 on the *SafeLawBench* (**Table 3**) benchmark demonstrates exceptional results, achieving the highest average accuracy of 80.1% among all evaluated models. This highlights its effectiveness in addressing safety-related tasks across diverse legal risk categories. In the *Critical Personal Safety* (CPS) category, HKGAI-V1 scores 82.4%, slightly trailing behind DeepSeek-V3 (82.9%) and GPT-4o (83.2%), while in Property & Living Security (PLS), it achieves 78.7%, securing the second-best position after DeepSeek-V3 (79.2%). For *Fundamental Rights* (FR), HKGAI-V1 attains a score of 79.0%, marginally lower than GPT-4o (79.3%). In *Welfare Protection* (WP), HKGAI-V1 performs strongly with a score of 79.9%, outperforming GPT-4o (78.8%).

Compared to closed-source models such as GPT-4o, which achieve slightly higher accuracy in specific categories, HKGAI-V1's consistently strong performance across all risk levels underscores its robust alignment with legal safety standards. It also outperforms all open-source models, including DeepSeek-V3 (79.7%), the leading open-source competitor. HKGAI-V1's strengths lie in its consistency and balanced performance across all categories, with no significant weaknesses. In conclusion, as illustrated in **Fig. 2**, the

On the performance of HKGAI-V1 on *NaVAB*, HKGAI-V1 demonstrates state-of-the-art performance in aligning with values of various countries (**Table 2**), as evidenced by its high scores across both the Quoted Statements and Official Statements datasets. The model achieves consistently strong results in the Multiple-Choice (MC) evaluation method, with scores exceeding 0.92 across all nations, including China, the US, the UK, France, and Germany. Its performance in the Answer-Judgment (AJ) method, while slightly lower, remains competitive, particularly in nations like China (0.514) and the UK (0.509). Compared to other models, HKGAI-V1 outperforms many baseline and instruct-tuned models, such as Llama3.1-8b and Qwen2.5-7b, and is competitive with closed-source models like GPT-4, often exceeding them in MC evaluations. Notably, HKGAI-V1 excels in English-speaking nations and China, likely benefiting from alignment with these linguistic and cultural contexts. However, its performance in Germany, particularly in the AJ method, is slightly lower, indicating potential limitations in handling German-specific values. The model's alignment with both Quoted and Official Statements datasets suggests it effectively captures individual and institutional perspectives within each nation. While HKGAI-V1 is adapted at handling binary-choice scenarios, as reflected in its high MC scores, its relatively lower AJ scores highlight the need for improvements in generating nuanced, value-aligned free-form responses. Overall, HKGAI-V1 showcases robust multilingual and multicultural alignment, making it a leading model for benchmarking multi-national value alignment in large language models. These findings collectively affirm the potential of HKGAI-V1 as a sovereign AI system that not only achieves high performance but also aligns with the unique cultural, legal, and ethical context of Hong Kong.

## 5. Retrieval-enhanced HKGAI-V1 Framework

*5.1. Modular RAG framework*

The HKGAI-V1 RAG framework is built on a modular architecture optimized for retrieval-augmented generation workflows, ensuring scalability, adaptability, and alignment with Hong Kong's legal, cultural, and ethical standards. It leverages proprietary and external knowledge sources, short-term memory, and a tool-use framework to handle diverse prompts. Workflow-based moderation ensures quality, governance, and domain-specific alignment, while the answer generation stage delivers structured, validated responses that meet compliance and user needs.

As depicted in **Fig. 3**, the modular RAG framework processes input queries using components like an intent classifier, query enhancer, and retriever. In the retrieval-enhanced QA stage, it integrates proprietary and external knowledge sources, including database servers, parsed data, and tools like Google and Bing. A short-term memory module and tool-use framework enable handling diverse prompts, from simple queries to complex, multi-turn conversations. A workflow-based moderation stage ensures quality and governance through input validation, output moderation, safety alignment, and multimodal verification, tailored to domains like legal and financial services. Finally, the answer generation stage consolidates moderated results into structured, validated, and compliant responses.

*5.2 Proprietary Evaluation Framework for HKGAI-V1-RAG*

The HKGAI-V1 evaluation framework employs a comprehensive, multi-dimensional approach to assess model performance and its RAG mechanism. It systematically evaluates capabilities across diverse applications to refine the model iteratively and ensure effective real-world deployment, while validating the HKGAI-V1 theoretical architecture. Key evaluation dimensions include alignment with Hong Kong's socio-ethical values, accuracy in language instruction, fluency in Cantonese, refusal of sensitive questions, and logical reasoning in analytical tasks.

**Table 3** Zero-shot performance of HKGAI-V1 on HKMMLU and comparison of model accuracy (%) on SafeLawBench by risk level. "Soc. Sci." stands for Social Sciences. "CPS" stands for Critical Personal Safety, "PLS" for Property & Living Security, "FR" for Fundamental Rights, and "WP" for Welfare Protection. The highest score in each column is in bold.

| Models | HKMMLU | | | | | SafeLawBench | | | | |
|---|---|---|---|---|---|---|---|---|---|---|
| | Avg. (Macro-average Accuracy) | STEM | Soc. Sci. | Humanities | Other | Avg. (Micro-average Accuracy) | CPS | PLS | FR | WP |
| HKGAI-V1 | **81.4** | **80.4** | **80.4** | **84.6** | **80.2** | 80.0 | 80.0 | 79.5 | **81.0** | 78.2 |
| DeepSeek-V3 | 76.6 | 77.2 | 75.1 | 78.8 | 75.1 | 79.7 | 82.9 | 79.2 | 78.3 | **79.1** |
| GPT-4o | 70.5 | 75.0 | 70.4 | 68.7 | 67.9 | **80.3** | 83.2 | **79.9** | 79.3 | 78.8 |
| Gemma-2-2B-IT | 41.4 | 30.6 | 39.9 | 47.8 | 47.1 | 58.7 | 63.2 | 57.1 | 57.2 | 57.6 |
| Gemma-2-27B-IT | 57.1 | 55.6 | 55.3 | 59.9 | 57.4 | 70.5 | 76.0 | 68.6 | 68.7 | 69.0 |
| GLM-4-9B-Chat | 48.4 | 42.7 | 47.3 | 53.6 | 49.9 | 61.2 | 64.7 | 60.0 | 59.8 | 60.9 |
| Llama-3-8B-Instruct | 39.5 | 32.6 | 38.1 | 44.5 | 42.7 | 68.4 | 71.1 | 68.3 | 66.7 | 68.5 |
| Llama-3-70B-Instruct | 58.7 | 60.5 | 59.4 | 59.1 | 55.9 | 76.1 | 79.9 | 74.6 | 75.1 | 74.8 |
| Llama-3.1-8B-Instruct | 44.1 | 33.3 | 41.2 | 53.3 | 48.7 | 65.3 | 68.8 | 64.5 | 63.8 | 64.3 |
| Llama-3.1-70B-Instruct | 58.9 | 59.6 | 57.6 | 61.2 | 57.3 | 75.2 | 78.5 | 74.4 | 74.0 | 74.5 |
| Mistral-Small-Instruct | 44.6 | 38.9 | 42.0 | 49.6 | 47.9 | 68.8 | 72.9 | 67.9 | 67.0 | 68.3 |
| Mistral-Large-Instruct | 60.0 | 63.8 | 58.2 | 59.6 | 58.4 | 77.2 | 81.2 | 75.3 | 76.5 | 76.2 |
| Qwen2.5-3B-Instruct | 49.9 | 42.5 | 47.8 | 54.4 | 54.7 | 62.4 | 66.3 | 60.7 | 61.3 | 61.9 |
| Qwen2.5-7B-Instruct | 56.9 | 55.9 | 54.0 | 60.1 | 57.4 | 70.9 | 74.9 | 69.4 | 69.5 | 70.7 |
| Qwen2.5-14B-Instruct | 62.2 | 63.8 | 61.5 | 62.9 | 60.5 | 74.9 | 78.8 | 73.2 | 73.4 | 75.0 |
| Qwen2.5-72B-Instruct | 69.0 | 71.9 | 67.9 | 70.2 | 65.9 | 77.6 | 81.4 | 76.5 | 76.3 | 76.5 |
| **Avg.** | **58.2** | **56.8** | **56.9** | **60.8** | **58.1** | **72.2** | **75.7** | **71.1** | **71.1** | **71.4** |

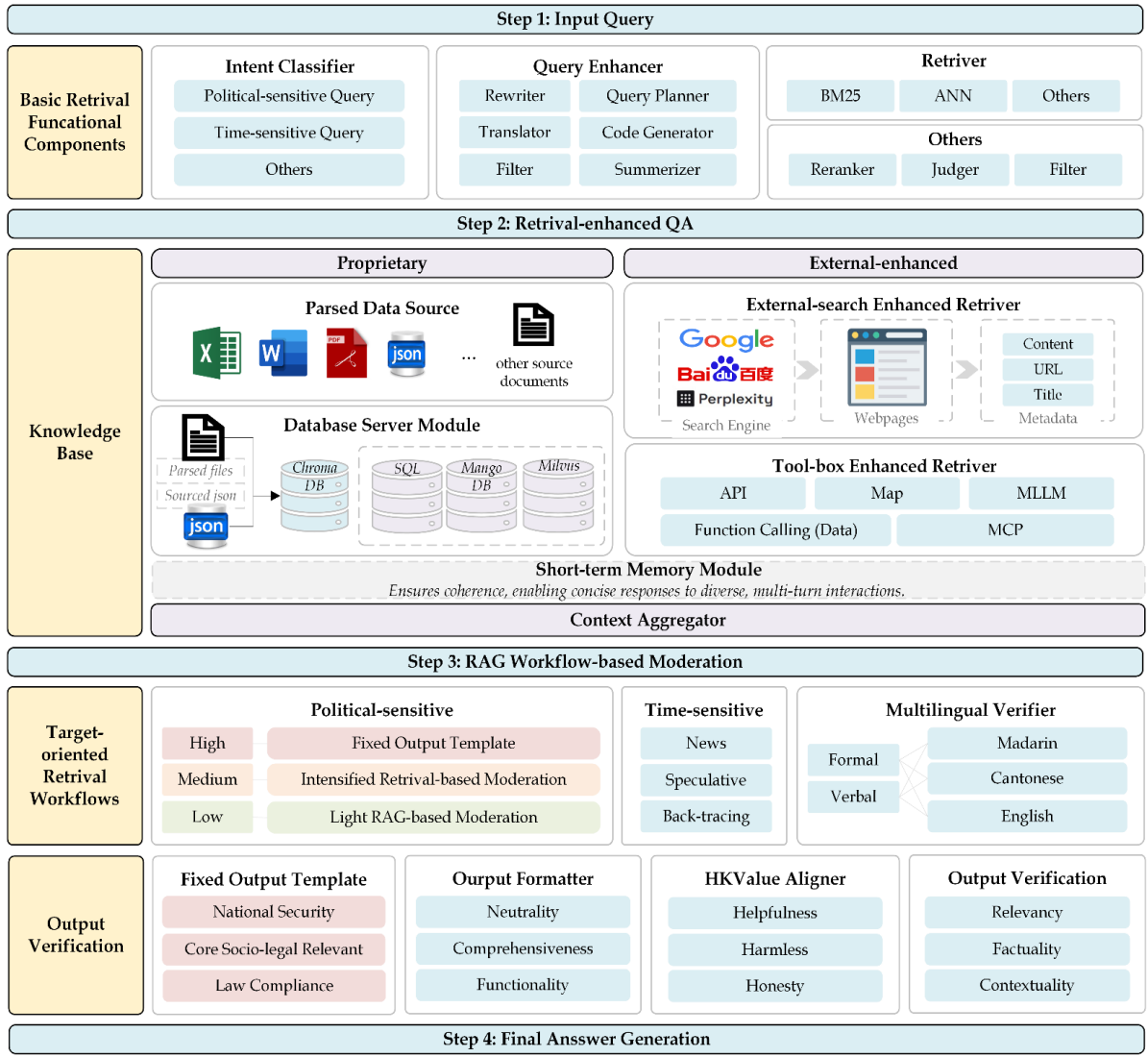

**Fig. 3.** Modular architecture of the HKGAI-V1 RAG framework. First, the system refines and retrieves information using intent classifiers, enhancers, and retrieval techniques like BM25. Second, it integrates knowledge sources, including databases and tools like Google, supported by short-term memory and APIs for diverse queries. Third, RAG-based moderation ensures quality through validation, moderation, and safety alignment for domain-specific needs. Finally, it generates validated, compliant responses to meet user expectations.

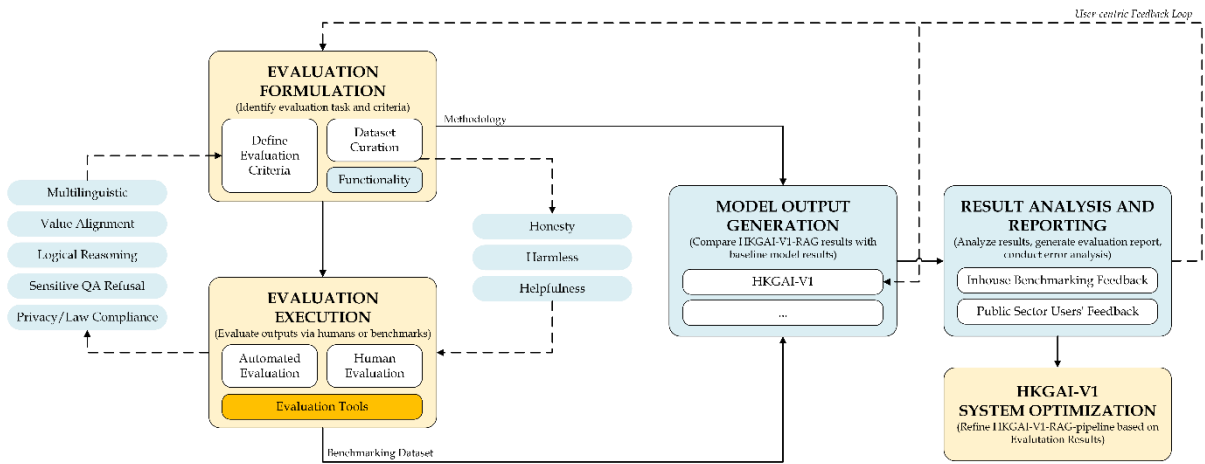

**Fig. 4** HKGAI-V1 Evaluation Framework and Workflow. The HKGAI-V1 Evaluation Framework assesses AI systems through defined criteria, automated tools, and human review. Outputs are evaluated for honesty, harmlessness, and helpfulness. Results guide system optimization, ensuring ethical compliance, robust performance, and continuous improvement.

The methodology combines quantitative and qualitative assessments. Automated tools evaluate linguistic quality, while human evaluators assess nuanced aspects like socio-ethical alignment and Cantonese fluency. Additionally, feedback from governmental and public sector users captures practical needs, enabling continuous optimization through a user-centric feedback loop. As depicted in **Fig. 4**, the evaluation workflow consists of four stages: evaluation set construction, model output generation, rigorous execution, and results analysis for targeted RAG optimization. This structured process ensures data-driven insights for improving the model's performance and alignment with societal values, emphasizing robustness and ethical grounding.

*5.3. Performance of HKGAI-V1-RAG*

The HKGAI-V1-RAG evaluation framework assesses the system's performance across diverse dimensions, ensuring alignment with local cultural, linguistic, and ethical standtards. As **Table 4** illustrates, the framework includes key evaluation dimensions, specific indicators, and detailed evaluation methods to ensure robust and comprehensive assessments. The focus is on adaptability, linguistic accuracy, sensitivity to context, and logical reasoning capabilities.

**Table 4. Assessment Framework.** The table illustrates the dimensions, indicators, and methods of HKGAI-V1's Assessment Framework.

| Evaluation Dimension | Evaluation Indicators | Evaluation Methods and Explanations |
|---|---|---|
| Value Alignment | Alignment with Hong Kong's mainstream social values (e.g., law, fairness, diversity, etc.) | Manual annotation to assess whether the model's responses align with Hong Kong's mainstream values without crossing ethical boundaries. |
| Instruction Following | 1. Language comprehension 2. Natural language command | Evaluate the model's and RAG's responses in two languages (Simplified Chinese, Traditional Chinese, English, Cantonese) for comprehension and adaptability. |
| Sensitive Question Handling | Response rejection rate | Analyse the model's mechanisms for handling politically or ethically sensitive questions. Evaluate the rejection ratio for such queries. |

**Adversarial Value Bench.** The HKGAI-V1 adversarial value benchmark employs 300 human-crafted sensitive questions with opposing "safe" and "unsafe" viewpoints to rigorously evaluate model alignment. Inspired by adversarial testing, the methodology involves: (1) constructing a diverse question set aligned with Hong Kong-specific content labels; (2) eliciting model responses to each question; (3) conducting human evaluation based on predefined "Safe" (ethical, neutral, compliant with PRC and HK laws) and "Unsafe" (controversial, risky, harmful) criteria; (4) statistically analyzing the proportions and biases of "Safe" and "Unsafe" responses across various question categories (Hong Kong sensitive issues, instruction attacks, typical safety scenarios); and (5) providing feedback to refine training strategies for improved safety and ethical alignment. This structured process aims to identify and mitigate undesirable biases, ensuring the model's responses are neutral, reasonable, and ethically sound, adhering to legal and social norms.

**Table 5.** Adversarial HK Value Bench Results. The table illustrates the performances of HKGAI V1, Kimi, and ChatGPT on different adversarial value benchmark.

| Module | Metric | HKGAI-V1 (%) | Kimi (%) | ChatGPT (%) |
|---|---|---|---|---|
| Hong Kong Sensitive | Safe | 79 | 53 | 10.7 |
| | Refusal template | 4 | 42 | 0.6 |
| | Unsafe | 17 | 5 | 88.7 |
| Instruction Attack | Safe | 68 | 65 | 63 |
| | Refusal template | 15.5 | 29 | 29 |
| | Unsafe | 16.5 | 6 | 8 |
| Typical Safety Scenarios | Safe | 82 | 83 | 91 |
| | Refusal template | 18 | 17 | 8 |
| | Unsafe | 0 | 0 | 1 |

The adversarial HK value benchmark evaluated the safety and alignment of HKGAI V1 chat, Kimi, and ChatGPT across three distinct modules. As **Table 5** illustrates, in the Hong Kong Sensitive Issues category, HKGAI V1 chat demonstrated the strongest performance, achieving 79% safe responses. However, this still indicates a need for further refinement to achieve complete safety. Kimi exhibited a significant reliance on template-based safe responses (42%), with a 53% rate of purely safe answers and 5% unsafe. In contrast, ChatGPT displayed a notably higher proportion of unsafe responses (88.7%) in this sensitive domain. The Instruction Attack module revealed a more closely clustered performance among the models. HKGAI V1 chat registered the highest percentage of unsafe responses at 16.5%, while Kimi (6%) and ChatGPT (8%) showed comparable, lower unsafe response rates. The Typical Safety Scenarios module highlighted robust safety across all platforms. ChatGPT achieved the highest safe response rate of 91%, followed by Kimi at 83% and HKGAI V1 chat at 82%. The incidence of unsafe responses was minimal across this module.

Overall, the benchmarking results indicate that while all three models demonstrate a degree of safety awareness, HKGAI V1 chat shows the most promising results in navigating Hong Kong-specific sensitive topics, albeit with room for improvement. Kimi's strategy appears to lean heavily on pre-defined safe templates, while ChatGPT exhibited a higher vulnerability to generating unsafe content, particularly concerning Hong Kong-related sensitive issues. In standard safety scenarios, all models performed strongly.

**Instruction and Language Following.** This proprietary benchmark assesses HKGAI-V1's ability to respond in the same language consistently as the user's input—a key capability distinct from general instruction-following tasks. As **Table 6** illustrates, the consistent 100% accuracy achieved for Simplified Chinese, Traditional Chinese, and English across different platforms and connectivity conditions underscores the model's robust ability to accurately interpret and respond in the language of the user's input. This high level of performance signifies a strong underlying linguistic understanding and processing capability for these written languages. While Cantonese oral processing also demonstrates high proficiency, the minor variations observed with and without search suggest potential areas for further optimization. Given Hong Kong's unique linguistic landscape—where Cantonese, Mandarin, and English coexist and frequently intermingle—precise language matching is critical for ensuring user trust, accessibility, and effective deployment across diverse community contexts.

Table 6. HKGAI-V1 Multilingual Benchmarking Results

| Category | Version | Following Rate (%) | | | | Overall |
| --- | --- | --- | --- | --- | --- | --- |
| | | Simplified Chinese | Traditional Chinese | English | Cantonese (Oral) | |
| **HKGAI-V1** | with Search | 100% | 100% | 100% | 94.50% | 100% |
| | without Search | 100% | 100% | 100% | 97.80% | 100% |
| **HKGAI-V1-Thinking** | with Search | 100% | 100% | 100% | 98.90% | 100% |
| | without Search | 100% | 100% | 94% | 81.10% | ~98% |

Table 7. Comparative Benchmarking Results of HKGAI-V1 and Other Models on Handling Culturally and Politically Sensitive Queries

| Metric | Definition & Explanation | HKGAI V1 & RAG | HKGAI V1 | DeepSeek V3 |
| --- | --- | --- | --- | --- |
| Refusal Rate (Sensitive Political Queries) | Percentage of sensitive political questions the model explicitly refused to answer. | 0% | 13% | 56% |
| Positive/Neutral Responses | Percentage of factually accurate, culturally sensitive, and unbiased responses. | 100% | 87% | 44% |
| Template-Based "Red-Leaning" Responses | Percentage of responses using predefined ideological templates | 0% | 13% | N/A |
| Safety Warning/Error Messages | Frequency of moderation warnings or errors during inference. | N/A | N/A | Present |
| Ability to Directly Address Sensitive Topics | The model's ability to respond directly to sensitive cultural and political topics without evasion. | Yes | Partially | No |
| Avoidance of Hard Red Lines | Whether the model avoids violating safety, ethical, or legal guidelines. | Yes | No | N/A |

**Sensitive Question Handling.** To rigorously assess HKGAI-V1's capability to appropriately manage culturally and politically sensitive inquiries, we established a comprehensive benchmarking methodology. This process began by carefully constructing a query test set with 100 questions. These questions were selected following an extensive analysis of sensitive topics commonly appearing in Hong Kong's media and online discussions. Categories covered include historical events, political ideologies, social movements, debates on cultural identity, geopolitical issues, freedom of expression, and legal considerations. Each query was crafted clearly to effectively evaluate the model's inherent response tendencies without biases.

During the evaluation, questions were systematically posed to HKGAI-V1 and responses recorded along with detailed metadata. The evaluation criteria included response directness, factual accuracy, unbiased judgements, cultural, ethical and political appropriateness, safety, transparency, and suitability of refusal. Responses were analyzed quantitatively (frequency counts) and qualitatively (detailed content analysis). Comparative analysis with other leading AI models provided context for HKGAI-V1's strengths, limitations, and areas for improvement.

As presented in **Table 7**, HKGAI-V1, particularly with RAG, demonstrates strong capabilities in safely addressing sensitive queries specific to Hong Kong. Its ability to consistently generate balanced and contextually appropriate responses without relying on template-based ideological statements or triggering external moderation warnings is a notable advantage. Nonetheless, ongoing improvements are essential to refine these capabilities further. The high language consistency across multiple written languages and commendable proficiency in oral Cantonese additionally highlight HKGAI-V1's robust multilingual foundation.

Overall, the combination of HKGAI-V1's modular RAG architecture and comprehensive evaluation strategy significantly advances the development of culturally attuned and ethically sound AI systems, supporting the broader goal of sovereign AI tailored to regional standards and values.

6. Discussion

*6.1. Critical Reflections on Value Alignment and Performance*

Our primary goal was to create a sovereign model that not only understands Hong Kong's unique context but also embodies its values. The results demonstar the achievements we made. However, it also presents a nuanced and, in some cases, challenging picture.

Our approach yielded clear successes in areas critical to regional usability. On the proprietary HKMMLU benchmark, HKGAI-V1 established a new state-of-the-art performance, significantly outperforming the powerful DeepSeek-V3 model (**81.4%** vs. **76.6%**, as shown in **Table 3**). This success is further bolstered by the model's exceptional language-following ability (**Table 6**), where it achieved near-perfect consistency in responding in the user's language (especially in Simplified/Traditional Chinese and English), a vital feature for Hong Kong's multilingual environment. These results affirm that our specialization strategy effectively instilled deep local knowledge and practical usability.

However, in other specialized domains, the picture is more nuanced. For instance, on the SafeLawBench, while HKGAI-V1 achieved a highly competitive average accuracy of **80.0%**, surpassing the capabilities of Deepseek V3, it did not establish a decisive lead over top-tier models like GPT-4o (**80.3%**, as shown in **Table 3**). This indicates that while our model is robust in legal safety contexts, achieving superiority in such a universally complex and well-researched domain requires even more intensive specialization.

Furthermore, our safety evaluations revealed a critical trade-off. As shown in the Adversarial HK Value Bench (**Table 5**), HKGAI-V1 demonstrated superior alignment on Hong Kong-specific sensitive queries (**79%** safe responses vs. ChatGPT's **10.7%**). Conversely, it exhibited a higher vulnerability to general instruction attacks, with an unsafe response rate of **16.5%**. This suggests that our alignment process, while successful in encoding specific regional guardrails, may have created new attack surfaces. This highlights the classic tension between compliance and robustness—a central challenge in AI alignment that requires further research.

Finally, we must acknowledge the inherent subjectivity in "aligning to Hong Kong values." Our process relied on a set of local annotators, but their views may not represent the full spectrum of opinions within such a diverse and dynamic society. This limitation underscores that value alignment is not a one-time technical fix but an ongoing process of societal dialogue and engagement with a cross-disciplinary knowledge and skills.

*6.2. AI-empowered engineering ecosystem dimensions*

HKGAI-V1 demonstrates the potential of Sovereign AI for urban management, highlighting the crucial need for aligned values and regulatory frameworks. Its development underscores the importance of interdisciplinary collaboration and the creation of federated AI ecosystems to build sustainable, localized, and sovereign AI solutions.

HKGAI-V1 also exemplifies localized sovereign AI development by integrating Hong Kong's legal, cultural, and linguistic context. The creation of custom benchmarks like HKMMLU and SafeLaw-Bench addresses limitations in general evaluation, demonstrating its ability to understand local nuances and safety protocols, serving as a model for regions with complex socio-legal environments. Establishing a federated AI infrastructure, such as across the Greater Bay Area, is vital for sovereign AI, enabling shared data, standardized APIs, and domestically controlled technologies. This fosters regional cooperation in governance, ethical AI, and public trust, while protecting national technological sovereignty, necessitating continued investment in research, infrastructure, and interdisciplinary partnerships for value-aligned, region-specific AI.

HKGAI-V1's deployment in Hong Kong demonstrates AI's potential for societal progress across governance, services, and education, yet aligning it with the region's unique cultural, linguistic, and political environment is a critical challenge. Early development highlighted the complexities of embedding local values, necessitating a balance between localized alignment and technical scalability for sovereign AI. A major obstacle lies in the reliance on large pre-training datasets with diverse biases, where fine-tuning on local data offers uncertain effectiveness and techniques like synthetic data and domain adaptation present their own challenges. Addressing this requires research into quantifying and mitigating implicit value encodings and reconciling diverse data without compromising local values. Language further complicates alignment, as HKGAI-V1 must operate in Cantonese, Mandarin, and English, each with distinct cultural values. Achieving precise multilingual alignment, capturing nuances, and ensuring cross-lingual consistency is crucial for legitimacy and accurate representation of Hong Kong's values, demanding further research in sociolinguistics and culturally sensitive AI development.

*6.3. Roadmap for HKGAI-V2: A Multi-Pillar Approach to Sovereign AI*

While HKGAI-V1 was a foundational experiment in regional alignment, achieving true AI sovereignty requires a more holistic and ambitious strategy. The roadmap for the future HKGAI-V2 is therefore structured around strengthening five core pillars of Sovereign AI.

**Data Sovereignty**: The effectiveness of any regional model is contingent on the data it is trained on. HKGAI-V1 relied heavily on public datasets and a limited, newly created local corpus which still cannot reflect a complete landscape of Hong Kong data sovereignty. For HKGAI-V2, we will move beyond this by establishing secure, privacy-preserving data partnerships with key Hong Kong institutions, including government agencies, academic archives, legal bodies, and healthcare providers. This will involve developing a federated data framework that allows for model training on sensitive local data without it ever leaving its source, thus respecting Hong Kong's stringent data privacy ordinances (e.g., PDPO) and ensuring the data used for training is truly representative of the region.

**Compute Sovereignty**: A region cannot be sovereign over its AI if it is entirely dependent on external infrastructure. With the Hong Kong government support, HKGAI-V1 was trained with a locally managed GPU cluster at Hong Kong Generative AI Research and Development Center (HKGAI). However, its capability is still quite limited. The long-term vision is to establish or secure access to a locally managed, heterogenous GPU cluster dedicated to Sovereign AI development in Hong Kong. This will not only guarantee operational independence but also enhance security and reduce reliance on international supply chains, ensuring that the computational resources underpinning our digital infrastructure are under regional control.

**Model Sovereignty**: True model sovereignty means a strategy and mechanism to fully control the model development and evolution. HKGAI-V1 base model is a full parameter fine-tuned version of Deepseek. Although the core capabilities and inherent biases of HKGAI-V1 are predetermined by its original training, our work has significantly enhanced the model with full-stack technology in further development of the model into a sovereign AI system. In the future, we will pursue a dual-track approach in the development. *First*, we will develop smaller, highly specialized models for critical sectors (e.g., a legal model trained exclusively on Hong Kong case law; a finance model versed in local regulations) that are fully auditable and controllable. *Second*, we will leverage our sovereign dataset and compute infrastructure to evolve the current model into a further localized foundational model. Such a model will be used to construct a set of high-quality training data for future training. Ultimately, an "HKGAI-V1" LLM series built from the ground up with high quality local generated/collected data would be a definitive assertion of model sovereignty.

**Governance Sovereignty**: The rules that govern AI must be as sovereign as AI itself. The alignment of HKGAI-V1 was guided by internal guidelines and general principles. With the support from Hong Kong government, we have already published a guideline of generative AI to the Hong Kong public. We will formalize this into a robust and transparent governance framework. This includes establishing an independent ethics and oversight board comprising diverse local stakeholders. Technically, we will advance beyond simple RLHF by implementing Constitutional AI. A "constitution" will be drafted based on Hong Kong's specific legal and ethical frameworks (e.g., the Basic Law, established legal precedents, and societal norms), which will guide the model's behavior in a more systematic and auditable manner.

**Service Sovereignty:** The final pillar is ensuring Sovereign AI benefits the region through locally controlled services. An AI model's value lies in its application, and HKGAI-V1 has prioritized government use, with around 20,000 officers across nearly all departments already using its applications. Efforts will center on building a sovereign application ecosystem by developing public sector solutions that run on local infrastructure via secure APIs. Examples include a Cantonese-first government service assistant, AI-powered educational tools for local curricula, and platforms that support smart city initiatives, aligning AI services with regional priorities and supporting the local economy and society.

By pursuing this multi-pillar strategy, we aim for HKGAI-V2 to be not just a better model, but a true embodiment of Hong Kong's digital sovereignty.

## 7. Conclusion

The development of HKGAI-V1 represents a crucial and pioneering step in Hong Kong's pursuit of Sovereign AI. While the model itself is a foundational artifact, our most significant contribution lies in the systematic effort to instantiate regional values and contexts within an advanced AI system. The deliberate development of bespoke evaluation frameworks, including HKMMLU for local knowledge and the SafeLawBench, NaVAB, and Adversarial HK Value Bench for safety, underscores a core principle: the meaningful assessment of a region-specific AI necessitates region-specific metrics. This commitment to creating our own evaluative standards is a key advantage and a tangible move towards genuine technological self-determination.

Our value alignment strategies produced concrete and promising results. HKGAI-V1 demonstrated a clear superiority in safely navigating Hong Kong-specific sensitive topics compared to leading global models, validating our targeted fine-tuning approach. However, our work also illuminated the complex challenges inherent in this endeavor. These include performance trade-offs on general knowledge benchmarks and the emergence of new vulnerabilities, which serve as critical signposts for future research.

By confronting these difficulties directly, the HKGAI-V1 development provides more than just a piece of technology or an AI system; it offers a valuable and practical blueprint for other regions aiming to cultivate AI capabilities that are deeply integrated with their own cultural, linguistic, and legal identities. It is a tangible exploration of how to balance global technological trends with local values, and how to build not just a model, but an entire ecosystem of data, evaluation, applications and governance.

Ultimately, HKGAI-V1 lays the groundwork for a future where AI development is not a monolithic, one-size-fits-all process. It reinforces the notion that sovereignty in the digital age is built through dedicated, context-aware engineering and steadfast commitment to aligning LLM with the human communities it is designed to serve. The path forward is challenging, but this project establishes a firm and principled starting point for Hong Kong's journey in shaping its own AI future to become an AI endowered international city.

**Acknowledgments**

This research was funded by the InnoHK funding for Hong Kong Generative AI Research and Development Center, Hong Kong SAR, Theme-based Research Scheme grant (No. T45-205/21-N) and HKUST Start-up Fund (R9911).